# Properties of Bayesian Belief Network Learning Algorithms

Remco R. Bouckaert – PhD student
Utrecht University
Department of Computer Science
P.O.Box 80.089 3508 TB Utrecht, The Netherlands
remco@cs.ruu.nl

## Abstract

In this paper the behavior of various belief network learning algorithms is studied. Selecting belief networks with certain minimallity properties turns out to be NP-hard, which justifies the use of search heuristics. Search heuristics based on the Bayesian measure of Cooper and Herskovits and a minimum description length (MDL) measure are compared with respect to their properties for both limiting and finite database sizes. It is shown that the MDL measure has more desirable properties than the Bayesian measure. Experimental results suggest that for learning probabilities of belief networks smoothing is helpful.
**Keywords:** Bayesian belief networks, learning, Bayesian approach, minimum description length principle, K2.

## 1 Introduction

The framework of Bayesian belief networks offers a mathematically sound formalism for representing uncertainty in knowledge-based systems. Efficient algorithms are associated with the formalism for making inferences with knowledge represented in belief networks, [Henrion, 1990; Lauritzen and Spiegelhalter, 1988; Pearl, 1988]. In addition, the framework has proved its practical worth over the last few years. However, constructing belief networks with the help of human experts is a time-consuming task. Since more and more large databases become available, automated learning algorithms can help shorten the build and test cycle of a belief network by suggesting an initial set-up. Therefore, learning belief networks from data is an important research issue.

In various fields of science, a lot of research effort has been spent on the design of methods for learning Bayesian belief networks from different perspectives such as computer science, statistics, and philosophy. One of the most promising to date is a Bayesian method proposed by Cooper and Herskovits, [Cooper and Herskovits, 1992]. However, methods based on the minimum description length principle are rapidly gaining popularity, [Bouckaert, 1993; Lam and Bacchus, 1994; Suzuki, 1993; Wedelin, 1993]. In this paper, the properties of several learning algorithms are investigated.

All these methods incorporate three elements; a quality measure for deciding which of a set of network structures is best, a search heuristic for suggesting a set of network structures to compare, and an estimation method for learning the probabilities in the network.

In the next section, we give a short introduction to terms and notations used in the remainder of this paper. Section 3 is devoted to the properties of quality measures. In Section 4, we consider search heuristics. In Section 5, we investigate learning probabilities for a Bayesian belief network. To compare various learning algorithms, we performed some experiments, the results of which are discussed in Section 6. We conclude with some final considerations and directions for further research.

## 2 Preliminaries

A *Bayesian belief network* $B$ over a set of variables $U$ is a pair $(B_S, B_P)$. $B_S$ is called the *network structure* of $B$ and takes the form of a directed acyclic graph with one node for each variable in $U$. For simplicity, we assume that the variables in $U$ are discrete. $B_P$ is a set of conditional probability tables; for every variable $x_i \in U$, $B_P$ contains a conditional probability table with parameters $P(x_i|\pi_i)$ that enumerates the probabilities of all values of $x_i$ given all combinations of values of the variables in its *parent-set* $\pi_i$ in the network structure $B_S$. The distribution represented by such a network $B$ is $P(U) = \prod_{x_i \in U} P(x_i|\pi_i)$, [Pearl, 1988].

In a network structure $B_S$, a *trail* is a path that does not consider the direction of the arcs. A *head-to-head node* in a trail is a node $e$ such that the sequence $x \to e \leftarrow y$ is part of the trail. A trail between two nodes $x$ and $y$ is *blocked* by a set of nodes $Z$ if at least one of the following two conditions hold:
• the trail contains a head-to-head node $e$ such that $e \notin Z$ and every descendant of $e$ is not in $Z$;
• the trail contains a node $e$ such that $e \in Z$ and $e$ is not a head-to-head node in the trail.

In a network structure $B_S$, let $X$, $Y$ and $Z$ be sets of nodes. We say that $X$ is *d-separated* from $Y$ given $Z$ if every trail between any node $x \in X$ and any node $y \in Y$ is blocked by $Z$.

For a joint probability distribution $P$ over a set of variables $U$, we call $X$ and $Y$ *conditionally independent* given $Z$, written $I(X, Z, Y)$, if $P(XY|Z) = P(X|Z)P(Y|Z)$ for all values of the variables in $XYZ$. $I(X, Z, Y)$ is called an *independency statement*.

A network structure $B_S$ is an *independency map* or *I-map* of a distribution $P$ if $X$ and $Y$ being d-separated by $Z$ in $B_S$ implies that $I(X, Z, Y)$ holds in $P$; $B_S$ is a *minimal I-map* of $P$ if no arc can be removed from $B_S$ without destroying its I-mappedness. $B_S$ is a *perfect map* or *P-map* of $P$ if it is an I-map and $I(X, Z, Y)$ holding in $P$ implies that $X$ and $Y$ are d-separated by $Z$ in $B_S$.

In discussing the various algorithms, we assume that cases in the database occur independently and, there are no cases that have variables with missing values.

## 3  Quality Measures

The task of learning Bayesian belief networks is twofold; learning the network structure $B_S$ and learning the set of probability tables $B_P$. The latter task will be addressed in a later section. Here, we investigate learning network structures.

For learning network structures, we use a quality measure. With such a measure a network structure that has a higher quality is preferred above a network structure with a lower quality. A heuristic search procedure is performed to select a network structure that has the highest quality among the considered network structures.

Cooper and Herskovits [Cooper and Herskovits, 1992] have proposed a quality measure based on a Bayesian approach in which they assume that no probability table $B_P$ is preferred for a given network structure before the database has been inspected.

Let $U$ be a set of $n$ discrete variables $x_i$, where each $x_i$ can take one of $r_i$ different values $v_{i1}, \ldots, v_{ir_i}$, $i = \{1, \ldots, n\}$ $r_i \geq 2$. Let $D$ be a database of $N$ cases where each case specifies a value for each variable in $U$. Let $B_S$ denote a network structure over the variables in $U$. For each $x_i$, let $\pi_i$ be the parent set of $x_i$ in $B_S$; let $w_{ij}$ denote the $j$th instantiation of $\pi_i$ relative to $D$, $j = 1, \ldots q_i$, $q_i \geq 1$. Now let $N_{ijk}$ be the number of cases in $D$ in which variable $x_i$ has the value $v_{ik}$ and $\pi_i$ is instantiated as $w_{ij}$, and let $N_{ij} = \sum_{k=1}^{r_i} N_{ijk}$. Let $P(B_S)$ be the probability of $B_S$ prior to observation of the database. Then, the probability of $B_S$ and the database $D$ is,

$$P(B_S, D) = P(B_S) \prod_{i=1}^{n} \prod_{j=1}^{q_i} \frac{(r_i - 1)!}{(N_{ij} + r_i - 1)!} \prod_{k=1}^{r_i} N_{ijk}!. \quad (1)$$

For a proof, we refer to [Cooper and Herskovits, 1992]. In practical implementations, generally the logarithm[1] of Equation (1) is taken; the result will be referred to as the *Bayesian measure* of the quality of a network structure.

Another measure for networks is the description length,

$$L(B_S, D) = \log P(B_S) - N \cdot H(B_S, D) - \frac{1}{2} k \cdot \log N \quad (2)$$

where $H(B_S, D)$ is defined as $\sum_{i=1}^{n} \sum_{j=1}^{q_i} \sum_{k=1}^{r_i} -\frac{N_{ijk}}{N} \log \frac{N_{ijk}}{N_{ij}}$ and $k$ as $\sum_{i=1}^{n}(r_i - 1) \prod_{x_j \in \pi_i} r_j$. It has been shown [Bouckaert, 1993] that (2) is approximately equal to the Bayesian measure except for error terms that are less than $O(1)$ with respect to $N$. Compared to the Bayesian measure, this measure has a more intuitive interpretation. When Formula (2) is interpreted as approximation of the Bayesian measure, the term $\log P(B_S)$ may seem superfluous since it is of the same order as the approximation errors. Still the term remains useful for incorporating prior information. For example, prior information on the existence and direction of arcs can be incorporated in this term. The term $-N \cdot H(B_S, D)$ is $-N$ times the entropy of a network structure $B_S$ and database $D$. Generally, this term increases as arcs are added to a network structure. In the last term, $k$ equals the number of independent probabilities that are needed to define all probability tables in $B_P$ for $B_S$. The term $-\frac{1}{2} k \cdot \log N$ thus models the cost of estimating these $k$ probabilities. Contrary to the entropy term, this term decreases when arcs are added to a network structure. When no prior information is available so $P(B_S)$ is equal for all network structures $B_S$, this measure assigns high quality to network structures that fit the database with as few arcs as possible. Therefore, Equation (2) will be referred to as the *minimum description length (MDL) measure* for the quality of a network structure.

### 3.1  Properties of Quality Measures

This section will be devoted to the investigation of the behavior of the Bayesian and MDL measures.

---

[1] All logarithms are to the base two.

We say that a network structure $B_S$ obeys a total ordering $<_U$ on the variables $U$ if for every arc $x_i \to x_j$ in $B_S$ $x_i <_U x_j$.

**Theorem 3.1** *Let $P_D$ be a positive distribution over a set of variables $U$. Let $D$ be a database with $N$ cases generated by $P_D$ where $N$ approximates infinity. Let $<_U$ be a total ordering on $U$. Let $B_S$ be the minimal I-map of $P_D$ that obeys $<_U$. Then, for any network structure $B_{S'}$ $B_{S'} \neq B_S$ that obeys $<_U$,*

$$M(B_S, D) \gg M(B_{S'}, D),$$

*where $M$ is either the Bayesian or the MDL measure.*

A proof of this theorem can be found in [Bouckaert, 1994]. The theorem states that for databases large enough, networks that are minimal I-maps that obey a particular ordering $<_U$ are overwhelmingly preferred over networks that are non-minimal I-maps. As a consequence, we have that if a P-map exists that obeys $<_U$ then this P-map will be overwhelmingly preferred.

It is interesting to investigate the properties of the quality measures when no ordering on the variables is provided. However, minimal I-maps are not unique when no ordering needs to be obeyed. Uniqueness of optimal structures is a desirable property since it helps deriving theoretical results and is important for deriving causal structure. Minimal I-maps need not even be equivalent, that is, represent the same set of independency statements. Consider for example Figure 1; if the structure on the left is a P-map of a distribution $P$, then both other structures on the right are minimal I-maps of $P$ marginalised over $b$. However, the upper structure represents $I(a, \emptyset, e)$ while the lower structure does not.

Also, the number of probabilities that need to be estimated for minimal I-maps need not be the same. This gives reason to distinguish between minimum and non-minimum structures. A *minimum I-map* $B_S$ of a distribution $P$ is a minimal I-map of $P$ such that in the belief network $B = (B_S, B_P)$ the least number of probabilities need to be specified in order to define all probability tables $B_P$. Consider once more Figure 1 and let all variables be binary; then, for both I-maps twelve probabilities need to be specified, one for $a$ one for $c$ and eight and two for $d$ and $e$ ($e$ and $d$) in the upper (lower) network structure. We have the following theorem for minimum I-maps.

**Theorem 3.2** *Let $P_D$ be a distribution over a set of variables $U$. Let the prior distribution over all network structures over $U$ be positive. Let $D$ be a database with $N$ cases generated by $P_D$ where $N$ approximates infinity. Let $B_S$ be a minimum I-map of $P_D$. Then, for any non-minimal I-map $B_{S'}$ of $P_D$ we have that,*

$$M(B_S, D) \gg M(B_{S'}, D),$$

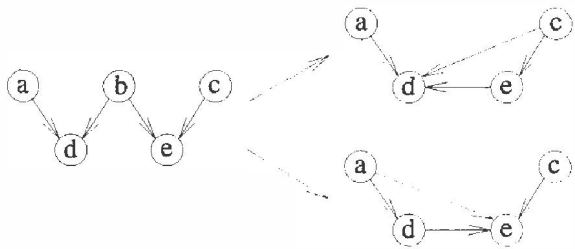

Figure 1: Example of different minimal I-maps.

*where $M$ is either the Bayesian or the MDL measure.*

The theorem suggests that for databases large enough, network structures that are minimum I-maps are preferred over network structures that are non-minimum I-maps. Note that contrary to Theorem 3.1 no positive distribution $P_D$ is demanded. The theorem also suggests that if a P-map exists for a distribution then a P-map is preferred over non P-maps for large databases; this follows from the property that all P-maps require the same number of probabilities and every I-map that is not a P-map requires more probabilities. A proof of this property can be found in [Bouckaert, 1994]. Note that also P-maps need not be unique; if $a \to b$ is a P-map then $a \leftarrow b$ is also a P-map.

We are interested in properties of network structures that are selected when a quality measure is used. The following theorem gives some insight.

**Theorem 3.3** *Let $D$ be a database with $N$ cases over $U$, $N \geq 10$. Let $B_S$ be a network structure over $U$, with at least one parent-set containing $\log N$ variables or more. If no prior information is available then, a network structure $B_{S'}$ exists such that $L(B_{S'}, D) > L(B_S, D)$.*

Again a proof can be found in [Bouckaert, 1994]. The theorem implies that search algorithms that use the MDL measure will not select network structures that contain parent-sets with more than $\log N$ parents. A similar result would be expected for the Bayesian measure. However, consider a database $D_i$ recursively defined by

$$D_1 = \begin{array}{cc} x_1 & y \\ 0 & 0 \\ 1 & 1 \end{array}$$

where the rows represent the individual cases and the columns the values of the variables; $D_n$ is constructed from $D_{n-1}$ by adding a column for an extra variable $x_n$ filled with 1s. Further, two identical cases are added where $x_i = 0$ ($1 \leq i \leq n$) and $y = (n+1) \bmod 2$. For example, Figure 2 shows how $D_7$ is build from $D_1$ up to $D_6$.

Figure 3 shows the network structure that scores highest with the Bayesian measure; all nodes $x_i$ have a node $x_{i-1}$ in their parent-set while $y$ has all nodes $x_1, \ldots, x_7$ in its parent-set. This network was found by brute force; the Bayesian measure was calculated for all 1.138.779.265 possible networks. Obviously, the parent-set of node $y$ contains more than $log(14)$ parents.

| $x_7$ | $x_6$ | $x_5$ | $x_4$ | $x_3$ | $x_2$ | $x_1$ | $y$ | |
|---|---|---|---|---|---|---|---|---|
| 0 | 0 | 0 | 0 | 0 | 0 | 0 | 0 | |
| 0 | 0 | 0 | 0 | 0 | 0 | 0 | 0 | $D_7$ |
| 1 | 0 | 0 | 0 | 0 | 0 | 0 | 1 | |
| 1 | 0 | 0 | 0 | 0 | 0 | 0 | 1 | $D_6$ |
| 1 | 1 | 0 | 0 | 0 | 0 | 0 | 0 | |
| 1 | 1 | 0 | 0 | 0 | 0 | 0 | 0 | $D_5$ |
| 1 | 1 | 1 | 0 | 0 | 0 | 0 | 1 | |
| 1 | 1 | 1 | 0 | 0 | 0 | 0 | 1 | $D_4$ |
| 1 | 1 | 1 | 1 | 0 | 0 | 0 | 0 | |
| 1 | 1 | 1 | 1 | 0 | 0 | 0 | 0 | $D_3$ |
| 1 | 1 | 1 | 1 | 1 | 0 | 0 | 1 | |
| 1 | 1 | 1 | 1 | 1 | 0 | 0 | 1 | $D_2$ |
| 1 | 1 | 1 | 1 | 1 | 1 | 0 | 0 | |
| 1 | 1 | 1 | 1 | 1 | 1 | 1 | 1 | $D_1$ |

Figure 2: Example of databases $D_i$ up to $D_7$.

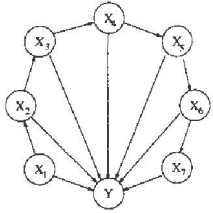

Figure 3: Most probable network for $D_7$

The same behavior arises for larger databases $D_j$ $j > 7$. The Bayesian measure will prefer parent-sets $\pi_i$ of $x_i$ if knowledge of the values of $\pi_i$ gives information of the distribution of $x_i$ while the number of states of the parent-set $q_i$ is not too large.

Let us consider the parent-sets of the best structure for $D_j$ according to the Bayesian measure. In $D_j$, nodes $x_i$ ($1 < i < j$) will have $x_{i-1}$ in their parent-set because when $x_{i-1}$ is 1, we find in the database that $x_i$ is 1. Furthermore, $x_{i+1}$ may be in $\pi_i$ because when $x_{i+1}$ is 0, we find in the database that $x_i$ is 0. The only variable that gives more information is $y$; $x_i$ is a function of $x_{i-1}$, $x_{i+1}$ and $y$ in the database. However, adding $y$ to the parent-set of $x_i$ would increase $q_i$ relatively much while the information obtained only applies to four cases in the database. For $x_j$ ($x_1$) a similar argument holds, except that $x_{j+1}$ ($x_0$) is non-existant and need not be considered.

What remains to be considered is node $y$. By an inductive argument, it can be shown that if $\pi_y$ is $x_k, \ldots, x_j$ $1 < k \leq j$ then taking $x_{k-1}, \ldots, x_j$ for $\pi_y$ results in a better score. Further, any parent-set of size $s$ gives exact information for at most as much cases for $y$ as $x_{j-s+1}, \ldots, x_j$ so these sets score worse. Therefore, the best scoring parent-set for $y$ is $x_1, \ldots, x_j$ in $D_j$. So, we have that in a database with $N$ cases, a network with parent-sets with $N/2$ variables can be assigned quality by the Bayesian measure.

So, while the asymptotic behavior of the Bayesian and MDL measure is the same, this is not the true for the practical case for which a finite database is available.

## 4 Search Algorithms

Recall that the purpose of an algorithm for learning network structure from data is to select a network structure with highest quality according to a given quality measure. The number of different network structures is more than exponential in the number of variables; for example, there are more than $4.2 \times 10^{18}$ network structures for 10 variables. The search space of possible structures for a learning algorithm is very large. We will show that the problem of finding a structure with minimality properties are NP-hard.

**Theorem 4.1** *Let $P$ be a given probability distribution over a set of variables $U$ and let an oracle be available that reveals wether an independency statement holds in $P$ or not. Then, the problem of finding an I-map of $P$ with a minimum number of arcs is NP-hard.*

A proof can be found in [Bouckaert, 1994]. From the theorem we have that no efficient algorithm exists for constructing a network structure with a minimal number of arcs, not even when we know the distribution that is to be represented. As a consequence we have that no minimum I-map can be found under the same conditions.

In [Cooper and Herskovits, 1992], Cooper and Herskovits proposed an algorithm, called K2. This algorithm departs from a given ordering on the variables. For simplicity, it is assumed that no prior information on the network is available, so the prior probability distribution over the network structure is uniform and can be ignored in comparing network structures. The algorithm is a greedy heuristic. For each variable $x_i$, the algorithm starts with an empty parent-set and tries to add variables to this set; a variable that is lower numbered than $x_i$ and maximally increases the quality is added to the parent-set $\pi_i$ of $x_i$. This process is repeated until adding such variables does not increase the quality anymore or the parent-set consists of all lower numbered variables.

In the algorithm, $m_i(\pi_i)$ represents the the contribution of variable $x_i$ with parent-set $\pi_i$ to the

---

**Algorithm K2**

Let the variables be ordered $x_1, \ldots, x_n$
for $i \in \{1, \ldots, n\}$ do          {initialize}
    $\pi_i \leftarrow \emptyset$
for $i \in \{2, \ldots, n\}$ do          {main loop}
    repeat
        select $y \in \{x_1 \ldots x_{i-1}\} \setminus \pi_i$ that
            maximizes $g = m_i(\pi_i \cup \{y\})$
        $\Delta \leftarrow g - m_i(\pi_i)$
        if $\Delta > 0$ then
            $\pi_i \leftarrow \pi_i \cup \{y\}$
    until $\Delta \leq 0$ or $\pi_i = \{x_1, \ldots, x_{i-1}\}$

---

**Algorithm B**

for $i \in \{1, \ldots, n\}$ do  $\pi_i \leftarrow \emptyset$    {initialize}
for $i \in \{1, \ldots, n\}, j \in \{1, \ldots, n\}$ do
    if $i \neq j$ then
        $A[i, j] \leftarrow m_i(\{x_j\}) - m_i(\emptyset)$
    else
        $A[i, j] \leftarrow -\infty$    {obstruct $x_i \to x_i$}
repeat                                   {main loop}
    select $i, j$, that maximize $A[i, j]$
    if $A[i, j] > 0$ then
        $\pi_i \leftarrow \pi_i \cup \{x_j\}$
        for $a \in Pred_i, b \in Desc_i$ do
            $A[a, b] \leftarrow -\infty$  {obstruct loops}
        for $k \leftarrow 1$ to $n$ do
            if $A[i, k] > -\infty$ then
                $A[i, k] \leftarrow m_i(\pi_i \cup \{x_k\}) - m_i(\pi_i)$
until $A[i, j] \leq 0$ or $\forall_{i,j} A[i, j] = -\infty$

---

measure used; we implicitly assume that we work with a database $D$, so $D$ is omitted in the notation. When the Bayesian measure is used, we have $m_i(\pi_i) = \log(\prod_{j=1}^{q_i} \frac{(r_i-1)!}{(N_{ij}+r_i-1)!} \prod_{k=1}^{r_i} N_{ijk}!)$ and when the MDL measure is used we have $m_i(\pi_i) = \sum_{j=1}^{q_i} \sum_{k=1}^{r_i} N_{ijk} \log \frac{N_{ijk}}{N_{ij}} - \frac{1}{2} q_i (r_i - 1) \log N$. Note that in theory, the change of the measure given the complete network need be computed. However, all terms other than the ones mentioned cancel out.

The main drawback of K2 is that it makes use of an ordering of the variables. If no reasonable ordering is available, a random ordering may be chosen which may be optimized in a post-processing step.

A learning algorithm that does not require an ordering of variables was suggested by [Buntine, 1991a]. This algorithm also uses a greedy heuristic. As K2, it starts with empty parent-sets for each node. In each step, an arc is added that does not introduce a cycle in the network abd maximally increases the quality of the network. Arcs are added until further addition does not increase the quality of the network or no candidate arcs are left.

An efficient implementation called algorithm B is shown below; $Pred_i$ is the set of indices of the predecessors of $x_i$ and $Desc_i$ is the set of indices of descendants of $x_i$ including $i$.

As the output of algorithm B need not be a minimal I-map, it is wise to have a post-processing step that optimizes the ordering of the variables as implied by the structure generated by algorithm B.

To conclude, we consider the complexity of algorithms K2 and B in terms of computations of $m_i(\pi_i)$. Both algorithms will calculate $m_i(\pi_i)$ at most $O(n^2 \cdot u)$ times where $u$ is an upper limit to the number of parents of a variable. Calculation of $m_i(\pi_i)$ takes at most $O(N \cdot r)$ calculations where $r = \max_{i \in \{1 \ldots n\}} r_i$.

## 5  Learning Distributions

The aim of learning Bayesian belief networks may be to obtain an approximation of a joint probability distribution over a set of variables for reasoning purposes. In the previous section, learning network structures was addressed. In this section, we will investigate the construction of the probability tables $B_P$ for a given network structure $B_S$.

The most obvious method to estimate the probabilities in $B_P$ is to use their expected value [Cooper and Herskovits, 1992],

$$\hat{\theta}_{ijk} = \frac{N_{ijk} + 1}{N_{ij} + r_i}, \qquad (3)$$

where $\hat{\theta}_{ijk}$ is the estimate for $P(x_i = v_{ik} | \pi_i = w_{ij})$. We will consider another method of learning probabilities in $B_P$, known as smoothing [Buntine, 1991b].

Suppose that for learning network structures the Bayesian measure is used. In the previous section, we considered selecting a *single* network structure with highest quality $P(B_S, D)$. Since $P(B_S, D)$ is fairly small, one could also consider selecting $m$ ($m > 1$) network structures ranking highest [Cooper and Herskovits, 1992]. The sum over $P(B_S, D)$ may be considerable larger. For the assessment of a probability $P(X|Y)$, this quantity may then be calculated in all $m$ Bayesian belief networks and their results can be weighted by the quality of the network structure. Since the MDL measure can be considered an approximation of the Bayesian measure, this procedure applies as well when using the MDL measure. However, it is computationally attractive to have only one network instead of $m$.

Fortunately, in special cases one single network can be used as the representation of $m$ different net-works. For a given ordering $<_U$, assume that for every variable $x_i$ a collection of parent-sets $\Pi_i$ of high quality is known and that all parent-sets obey $<_U$. One may expect that the set of Bayesian belief networks $S$ with structures $\{B_S | i \in \{1, \ldots, n\}, \pi_i \in \Pi_i\}$ will represent a large part of the quality. In [Buntine, 1991b], it was suggested that the set $S$ can be represented by one single Bayesian belief network $B'$ with network structure $B_{S'}$ with,

$$\pi'_i = \cup_{\pi_i \in \Pi_i} \pi_i,$$

for all variables $x_i$; $B_P$ specifies for $P(x_i = v_{ik} | \pi'_i = w'_{ij})$ the weighted sum,

$$\alpha_i \sum_{\pi_i \in \Pi_i} 2^{m_i(\pi_i)} \hat{\theta}^{\pi_i}_{ijk},$$

where $\alpha_i$ is a normalizing constant equal to $1/\sum_{\pi_i \in \Pi_i} 2^{m_i(\pi_i)}$; $\hat{\theta}^{\pi_i}_{ijk}$ is the estimated probability for the network where $x_i$ has parent-set $\pi_i$ as in (3).

The problem in applying this approach is to select collections of parent-sets of high quality. We observe that for efficient propagation of evidence in Bayesian belief networks, it is desired that the parent-sets are as small as possible. Therefore, the parent-sets need preferably have a large overlap. Taking a network structure generated by either algorithm K2 or B as final structures and letting $\Pi_i$ consist of the set of subsets of $\pi_i$ seems a reasonable procedure; it is expected that $\Pi_i$ contains high quality parent-sets and that $\pi_i$ is not very large, at least when the MDL measure is used. Subsets need to be determined and qualities of the parent-sets need to be calculated.

A more efficient solution is to incorporate the estimation of the probabilities into the search algorithm for learning network structure. The basic idea is that every time an arc $x_j \to x_i$ is added to the network structure to update the probability table of $x_i$. So, the set $\Pi_i$ consists of the subsets of the final parent-set of $x_i$ that once were parent-sets at some stage in the search algorithm. Note that the quality of these parent-sets need not be recalculated because they are directly available in the algorithm. The only extra administration needed is maintaining the sum of weights of parent-sets so far. Details can be found in the algorithm, called weighted K2, shown below. Of course, a similar approach can be applied to algorithm B and we will call this algorithm weighted B.

## 6  Experimental Results

To gain some insight in the performance of the different learning methods discussed in the previous sections, some experiments were performed. We have generated ten Bayesian belief networks with randomly chosen connected structures comprising ten binary variables; these networks were used as gold standard for comparing the networks yielded by the

---

**Algorithm weighted K2**

Let the variables be ordered $x_1, \ldots, x_n$
for $i \in \{1, \ldots, n\}$ do           {initialize}
   $\pi_i \leftarrow \emptyset$
   $w[i] \leftarrow \exp(m_i(\pi_i))$
   for $k \leftarrow 1$ to $r_i$ do $P_{i1k} \leftarrow \frac{N_{ik}+1}{N+r_i}$
for $i \in \{2, \ldots, n\}$ do           {main loop}
   repeat
      select $y \in \{x_1, \ldots, x_{i-1}\} \setminus \pi_i$ that
         maximizes $g = m_i(\pi_i \cup \{y\})$
      $\Delta \leftarrow g - m_i(\pi_i)$
      if $\Delta > 0$ then
         for $k \in \{1, \ldots, r_i\}, j \in \{1, \ldots, m_i\}$ do
            $P_{ijk} \leftarrow w[i].p(i, j, k) + e^g \frac{N_{ijk}+1}{N_{ij}+r_i}$
         $w[i] \leftarrow w[i] + e^g$
         $\pi_i \leftarrow \pi_i \cup \{y\}$
   until $\Delta \leq 0$ or $\pi_i = \{x_1, \ldots, x_{i-1}\}$
normalize $P_{ijk}$

---

various learning algorithms with. From the generated belief networks, databases of 100, 200, 300, 400, and 500 cases were constructed using stochastic simulatoin. These databases were used as input to the algorithms K2 and B; using the network structures yielded by the algorithms, a direct estimate of the required probabilities was made by applying Formula (3). Likewise, the databases were used as input to the algorithms weighted K2 and weighted B. The four algorithms were executed both with the Bayesian and with the MDL measure. For the ordering on the variables required by K2 and weighted K2, the ordering that was used to construct the original network was provided.

To analyze the performances of the different algorithms, we used the *divergence* or *cross entropy* measure defined by,

$$\sum_U P(U) . \log \frac{P(U)}{\hat{P}(U)}$$

where $\sum_U$ denotes the summation over all combinations of values of the variables in $U$, $P(U)$ is the probability of an instantiation of $U$ in the original network, and $\hat{P}(U)$ is the same probability in the estimated network. The divergence is a measure for how well the learning algorithms can capture the probability distribution that generated the database. If one is interested in learning the causal structure instead of the probability distribution, the number of erroneously placed arcs and edges in the embedded undirected graph would be a better measure. Experiments have shown that the both the Bayesian measure and MDL measure perform about equally well with K2 when the number of erroneously placed arcs is considered [Bouckaert, 1993]. However, with the MDL measure more arcs are omitted than with the Bayesian measure.

| | Average divergence | | | | | | | | |
|---|---|---|---|---|---|---|---|---|---|
| | Bayes measure | | | | MDL measure | | | | aver. |
| | Direct | | Weighted | | Direct | | Weighted | | |
| $N$ | K2 | B | K2 | B | K2 | B | K2 | B | |
| 100 | 0.973 | 1.253 | 0.976 | 1.255 | 0.753 | 0.929 | 0.652 | 0.870 | 0.766 |
| 200 | 1.062 | 1.040 | 0.950 | 1.034 | 0.693 | 0.661 | 0.689 | 0.669 | 0.680 |
| 300 | 0.960 | 0.995 | 0.920 | 0.916 | 0.530 | 0.440 | 0.583 | 0.522 | 0.587 |
| 400 | 1.084 | 1.161 | 1.061 | 1.123 | 0.740 | 0.527 | 0.689 | 0.506 | 0.689 |
| 500 | 0.481 | 0.440 | 0.477 | 0.460 | 0.315 | 0.223 | 0.517 | 0.220 | 0.313 |
| aver. | 0.912 | 0.978 | 0.877 | 0.957 | 0.606 | 0.556 | 0.626 | 0.557 | |
| | Variance of divergence | | | | | | | | |
| 100 | 0.981 | 0.741 | 0.561 | 0.562 | 1.160 | 0.763 | 0.621 | 0.597 | 0.599 |
| 200 | 0.546 | 0.634 | 0.501 | 0.662 | 0.354 | 0.271 | 0.343 | 0.258 | 0.357 |
| 300 | 0.276 | 0.306 | 0.186 | 0.164 | 0.286 | 0.109 | 0.276 | 0.124 | 0.173 |
| 400 | 0.549 | 0.656 | 0.457 | 0.599 | 0.623 | 0.400 | 0.558 | 0.339 | 0.418 |
| 500 | 0.147 | 0.074 | 0.142 | 0.071 | 0.111 | 0.059 | 0.242 | 0.057 | 0.090 |
| aver. | 0.500 | 0.482 | 0.369 | 0.411 | 0.507 | 0.320 | 0.408 | 0.275 | |

Table 1: Test results.

Table 1 shows the results of our experiments; each entry shows the divergence averaged over the ten databases of size specified in the first column. The last column specifies the averages over the various rows and the last row specifies the averages over the columns.

Surprisingly, networks reconstructed with the MDL measure on average have a considerably lower divergence than those reconstructed with the Bayesian measure; this is independent of the search heuristic used and the method of probability estimation. The reason for this property is that with the Bayesian measure, networks with many extra arcs were found compared to the original networks, due to the property described at the end of Section 3. It is evident that for variables with additional parents, more probabilities needed to be estimated. So, more errors were introduced in the probability tables than for networks constructed with the MDL measure, resulting in the higher divergence.

Weighted estimates resulted in better distributions for networks constructed with the Bayesian measure for a similar reason. When the MDL measure was used, however, the estimation method did not influence the average divergence. Yet, the variance of the divergence was reduced considerably. The test results suggest that it is useful to use a weighted estimate when only one network structure is required.

Larger databases resulted in lower average divergence, as was expected.

The choice of the search heuristic did not seem to have a dramatic effect on the divergence; sometimes algorithm K2 resulted in a slightly better network, and other times algorithm B gave a better result. However, this is under the condition that K2 is provided with a good ordering on the variables. The original ordering is not necessarily the best ordering on the variables if the database is finite.

## 7 Conclusion

We have investigated the influence of quality measures, search heuristics, and estimation methods on learning Bayesian belief networks.

We have shown that the Bayesian measure and the minimum description length (MDL) have the same properties for infinite databases. However, these measures differ on finite databases; for the MDL measure the sizes of parent-sets in selected network structures is smaller than the logarithm of the database size while for the Bayesian measure a parent-set may be as large as half the database size.

These experiments have indicated that the MDL measure seems to result in better network structures than the Bayesian measure. In my opinion this is due to the larger sensitivity of the Bayesian measure being more sensitive to coincidental correlations implied by the data than the MDL measure. As a consequence, the Bayesian measure will prefer network structures with more arcs over simpler network structures. The experiments suggest further that weighting methods for learning the probabilities of a Bayesian belief network results in better distributions than when the probability tables are estimated directly. This is most apparent when the Bayesian approach is used to select a network structure.

In this paper, it is shown that it is NP-hard to select a network structure with a minimum number of arcs that is an I-map of a given distribution. Therefore, it may not be expected that efficient algorithms exist for learning network structures from data. Further research should aim at finding efficient heuristics for this task. We believe that it is necessary and use-ful to develop a post-processing algorithm that optimizes network structures yielded by currently known search heuristics.

### Acknowledgements

I thank Linda van der Gaag, the anonymous referees and Marek Druzdzel for their helpful comments that improved the presentation of the paper considerably.